\documentclass{ecai} 

\usepackage{graphicx}
\usepackage{subcaption}
\usepackage{comment}
\usepackage{xcolor} 
\usepackage[nomessages]{fp} 
\usepackage{booktabs} 
\usepackage{multirow} 
\usepackage{url}
\usepackage{framed}
\usepackage{enumitem}
\usepackage{array}
\usepackage{tabularx}
\usepackage{balance}
\usepackage{amsmath} 
\usepackage{siunitx} 
\usepackage{subcaption}
\usepackage{adjustbox}
\usepackage[flushleft]{threeparttable}
\usepackage{setspace} 
\usepackage{caption} 
\usepackage{soul,color}
\usepackage{xspace}
\usepackage{amsmath}
\usepackage{arydshln}
\usepackage{lipsum}  
\usepackage{booktabs}
\usepackage{mdframed} 

\soulregister\cite7
\soulregister\ref7
\soulregister\pageref7
\setstcolor{red}
\setul{}{.2ex}

\usepackage{hyperref}
\hypersetup{
    colorlinks=true,
    linkcolor=blue,
    citecolor=blue,
    urlcolor=blue
}

\begin{document}
\begin{frontmatter}
\paperid{9556} 

\title{On Robustness and Reliability of \\Benchmark-Based Evaluation of LLMs}


%
\author{\fnms{Riccardo}~\snm{Lunardi}}
\author{\fnms{Vincenzo}~\snm{Della Mea}}
\author{\fnms{Stefano}~\snm{Mizzaro}}
\author{{\fnms{Kevin}~\snm{Roitero}$^{\text{\footnotesize{*}}}$\thanks{All authors are corresponding authors. Email: \{name.surname\}@uniud.it}}}
\address{University of Udine, Italy}

\begin{abstract}
Large Language Models (LLMs) effectiveness is usually evaluated by means of benchmarks such as MMLU, ARC-C, or HellaSwag, where questions are presented in their original wording, thus in a fixed, standardized format. However, real-world applications involve linguistic variability, requiring models to maintain their effectiveness across diverse rewordings of the same question or query. In this study, we systematically assess the robustness of LLMs to paraphrased benchmark questions and investigate whether benchmark-based evaluations provide a reliable measure of model capabilities. 

We systematically generate various paraphrases of all the questions across six different common benchmarks, and measure the resulting variations in effectiveness of 34 state-of-the-art LLMs, of different size and effectiveness. Our findings reveal that while LLM rankings remain relatively stable across paraphrased inputs, absolute effectiveness scores change, and  decline significantly. This suggests that LLMs struggle with linguistic variability, raising concerns about their generalization abilities and evaluation methodologies. Furthermore, the observed performance drop challenges the reliability of benchmark-based evaluations, indicating that high benchmark scores may not fully capture a model's robustness to real-world input variations. We discuss the implications of these findings for LLM evaluation methodologies, emphasizing the need for robustness-aware benchmarks that better reflect practical deployment scenarios.
\end{abstract}
\end{frontmatter}

\section{Introduction}

Large Language Models (LLMs) capabilities have rapidly advanced in recent years, demonstrating state-of-the-art performance across a variety of natural language processing tasks. Their effectiveness is primarily assessed through standardized benchmark evaluations, such as the Massive Multitask Language Understanding (MMLU) \cite{hendryckstest2021}, Winogrande \cite{10.1145/3474381}, and HellaSwag \cite{zellers-etal-2019-hellaswag}. These benchmarks provide a structured and widely accepted framework for comparing models, ensuring consistency in evaluation, in a multiple-choice question answering setting. However, questions are presented in a fixed, standardized wording and this controlled setting does not fully account for the variability inherent in real-world interactions, where users may express the same intent in multiple ways.

A crucial open question is whether LLMs evaluated in these rigid benchmark settings exhibit robustness to linguistic variation. Recent studies have shown that LLMs are highly sensitive to changes in prompt formatting and minor alterations in wording, often leading to significant performance fluctuations \cite{sclar2024quantifying, zhao-etal-2024-improving}. This phenomenon raises concerns about \emph{model robustness}, i.e., the ability of LLMs to generalize across semantically equivalent yet reworded inputs. If a model's effectiveness drops drastically when faced with paraphrased questions, its real-world applicability and generalization capabilities may be overestimated. 

At the same time, these observations challenge the \emph{reliability of benchmark-based evaluations}. If benchmark scores are highly dependent on the specific phrasing of questions, do they truly reflect a model's reasoning ability, or do they merely capture performance on a particular wording of the task? While prior research has explored the impact of prompt variations on LLM outputs \cite{10.1145/3627673.3680002}, less attention has been given to the extent to which benchmark evaluations remain \emph{stable and reliable} when test questions have systematic rewording.

In this study, we investigate both of these fundamental issues. First, we assess the \emph{robustness} of LLMs by systematically paraphrasing benchmark questions and measuring the resulting changes in performance. Second, we evaluate the \emph{reliability} of benchmark-based evaluations by analyzing whether the relative ranking of models remains stable across paraphrased inputs. By introducing controlled linguistic and syntactic variations, our study provides a direct test of models' ability to generalize beyond the specific phrasings seen during training. This allows us to assess whether current benchmark methodologies truly capture a model’s underlying reasoning capabilities or whether they merely reflect performance on narrowly framed tasks, ultimately risking an overstatement of model effectiveness.
We address the following research questions:
\begin{enumerate}[label={RQ\arabic*},leftmargin=*]
    \item \label{rq-reliable} Are benchmark-based evaluations \emph{reliable}? Do LLM evaluation results change significantly when benchmark questions are replaced by simple paraphrases?
    \item \label{rq-robust} Are LLMs \emph{robust} to question paraphrases? Does question paraphrasing decrease LLM effectiveness, revealing limitations in their generalization capabilities?
\end{enumerate}

Our findings reveal that while the rankings of LLMs remain relatively stable, their absolute performance drops significantly when questions are paraphrased. This suggests that while benchmarks may provide a reasonable comparative measure of models, they overestimate absolute performance and generalization abilities. 
All data and code, including the complete set of questions, paraphrases, model predictions, and high-resolution plots, are available at \url{https://osf.io/jq56f/?view_only=aa6d0c67db7e47c08cff3b3e1a610764}.

\section{Background and Related Work}
Effectiveness evaluation is a fundamental problem across multiple disciplines, including Information Retrieval (IR), Natural Language Processing (NLP), and Artificial Intelligence (AI). Reliable evaluation frameworks are essential for ensuring that progress in the field reflects genuine improvements rather than artifacts of specific test conditions. 
Effectiveness evaluation has a long tradition especially in the IR field, where a rich body of research has explored how to design evaluations that are both efficient and reliable, particularly when human annotations are expensive or limited \cite{harman2011information,INR-009,sparck1976information}. Studies have proposed methods to reduce the number of relevance judgments without compromising evaluation validity \cite{10.1145/1148170.1148219, 10.1145/2808194.2809470}, analyzed optimal topic set sizes for stable system comparisons \cite{10.1145/1629096.1629099}, and investigated model ranking stability under incomplete assessments \cite{voorhees2001philosophy}. These works collectively emphasize that relative rankings of retrieval systems tend to remain stable even under noisy or incomplete assessments \cite{10.1145/1390334.1390447,voorhees1998variations,10.1145/383952.383963}. Recent work has also raised concerns about the long-term reliability of test collections, showing that even if system rankings remain stable, benchmarks may lose discriminative power over time as models overfit to narrow interpretations of relevance, effectively ``expiring'' in their utility \cite{parry2025variations}.

Insights from decades of IR evaluation research can be leveraged in the context of LLM benchmarking, where structured evaluations have become the de-facto standard for measuring model capabilities. 
While benchmarks are widely used to track progress and compare model performance, their quality varies significantly, and many commonly used benchmarks suffer from issues such as poor design, limited replicability, or lack of statistical rigor \cite{reuel2024betterbench}. Benchmarks such as MMLU \cite{hendryckstest2021}, ARC-C \cite{clark2018think}, and HellaSwag \cite{zellers-etal-2019-hellaswag} provide standardized datasets that enable reproducible model comparisons. However, these benchmarks assume a rigidly controlled testing environment in which questions are presented in a fixed wording and format. This does not reflect real-world applications, where linguistic variability is the norm \cite{doi:10.1126/science.adj5957,mitchell2023abstraction,stevenson2024can}. In fact, humans may naturally phrase the same question in multiple ways depending on context, intent, or background knowledge, making it essential to assess whether models can maintain their effectiveness across diverse rewordings.

Recent studies have shown that LLMs are highly sensitive to prompt variations, with minor changes in wording, structure, or formatting leading to substantial performance shifts \cite{sclar2024quantifying, zhao-etal-2024-improving}. While prompt engineering techniques have investigated how to optimize LLM outputs by refining input phrasing \cite{wei2021finetuned}, this does not necessarily indicate robustness of models. Instead, it often exposes an over-reliance on surface-level patterns, where high benchmark scores may be driven by implicit dataset artifacts rather than genuine reasoning abilities \cite{10.1145/3627673.3680002}. This raises the question of whether LLM evaluations conducted under fixed benchmark conditions truly reflect a model's generalization capacity, or if they merely measure performance on a narrow set of specific phrasings.

Another relevant line of research focuses on Benchmark Agreement Testing (BAT),
which investigates whether different benchmarks yield consistent rankings of models. While there is evidence that rankings are preserved despite fluctuations in absolute scores, less attention has been given to whether the rankings remain stable when test questions themselves are paraphrased. If LLM rankings are highly sensitive to minor linguistic changes, this would suggest that benchmark-based evaluations may not fully capture the robustness of LLMs to real-world input variability.

Beyond static benchmarks, alternative evaluation methodologies have been proposed, such as using LLMs themselves as judges to assess model outputs \cite{kim2023prometheus,zheng2023judging,zhu2023judgelm}. While LLM-based evaluation provides scalability and cost-effectiveness, it introduces potential biases, particularly in favor of models that share similar architectures or training data distributions. Human evaluation remains the gold standard for assessing open-ended tasks, as seen in large-scale efforts such as ChatBot Arena \cite{zheng2023judging}, which aggregates human preferences to compute model rankings. However, human evaluation is costly, subject to annotator variability, and difficult to scale across extensive benchmark datasets. These limitations reinforce the continued reliance on structured benchmarks, despite their constraints.

The question of whether benchmark-based evaluations reliably reflect model robustness remains largely unexplored. While existing studies have investigated the stability of model rankings across different datasets or across answers \cite{salido2025none}, the impact of systematic question paraphrasing on LLM performance and benchmark reliability remains an open issue. Understanding whether LLMs maintain effectiveness across linguistic variations is crucial for ensuring that benchmark scores translate to real-world applicability. 
We systematically examine the effects of paraphrased inputs on LLM evaluation, contributing to a broader discussion on the limitations of current benchmarking methodologies and on the need for robustness-aware evaluation frameworks.

\section{Methodology}
We detail the benchmarks, the LLMs, the paraphrases generation method, and the prompts that we use in our experiments.

\subsection{Benchmarks}
We leverage six well-established benchmarks covering diverse reasoning and knowledge domains. 

\begin{description}
    \item[ARC-C.] The AI2 Reasoning Challenge (ARC-C) \cite{clark2018think} includes science questions that require deeper logical reasoning beyond simple fact retrieval. The dataset is explicitly constructed to be challenging for models relying on pattern recognition, demanding causal reasoning, scientific knowledge application, and inference abilities.
    
    
    \item[HellaSwag.] HellaSwag \cite{zellers-etal-2019-hellaswag} assesses commonsense reasoning by presenting a context followed by four possible continuations, one of which is the correct next sentence. Incorrect choices are adversarially generated to appear plausible, forcing models to rely on deeper narrative understanding rather than surface-level cues.
    
    \item[MMLU.] The Massive Multitask Language Understanding (MMLU) benchmark \cite{hendryckstest2021} evaluates LLMs across 57 diverse subjects, including history, mathematics, law, and computer science. It consists of questions designed to assess both factual recall and reasoning skills. MMLU is widely used due to its broad coverage and structured format.
    
    \item[OpenBookQA.] OpenBookQA \cite{mihaylov-etal-2018-suit} tests an LLM’s ability to apply elementary science knowledge beyond direct memorization. It consists of questions grounded in a curated set of science facts, requiring models to reason over both the provided knowledge and external commonsense information.
    
    \item[RACE.] RACE \cite{lai-etal-2017-race} is a reading comprehension dataset derived from English language exams for middle-high school students. Questions require to extract implicit meanings, summarize main ideas, and perform higher-level inference. Its diverse passage types and complex reasoning tasks make it a challenging benchmark.
    
    \item[SciQ.] SciQ \cite{welbl-etal-2017-crowdsourcing} covers physics, chemistry, biology, and earth sciences. Each question is paired with four answers and a supporting paragraph, allowing for retrieval-augmented evaluation. It assesses factual knowledge and ability to leverage explanations.
\end{description}
Each question has four possible answers, and only one of them is the correct choice. Table~\ref{tab:dataset} reports key statistics of the datasets. 
For the considered benchmarks, we paraphrased the full test split along with an additional sample of training instances, with sample size varying according to benchmark size: 5,000 examples for HellaSwag, MMLU, and RACE (larger benchmarks), 1,119 for ARC-C, 1000 for SciQ, and 500 for OpenBookQA (smaller ones). Additionally, for ARC-C, 13 questions were excluded as they contained 3 or 5 answer options instead of the expected 4.
We observed qualitatively similar patterns of variability across both benchmark test and training instances, in terms of how model responses changed across paraphrases. This held consistently across different benchmarks and model families. Given this, we present results over both test and sampled training data, treating them uniformly in our evaluation.

\begin{table}[tb]
    \centering
    \caption{Benchmarks statistics. The release date is presumed based on the first public appearance of the dataset.
    }
    \label{tab:dataset}
    \resizebox{\columnwidth}{!}{%
    \begin{tabular}{lrrrr}
    \toprule
    \textbf{Benchmark} & \textbf{Considered} & \textbf{Paraphrasable} & \textbf{Number of} & \textbf{Release} \\
    \textbf{Name} & \textbf{Questions} & \textbf{Questions} & \textbf{Paraphrases} & \textbf{Date}\\
    \midrule
    ARC-C         & 2,566 & 2,566  & 12,830  & 2018-03-14 \\
    HellaSwag     & 15,011 & 15,011 & 75,050  & 2019-05-29 \\
    MMLU          & 18,957 & 18,955 & 94,775  & 2020-09-10 \\
    OpenBookQA    & 1,487 & 1,487  & 7,435   & 2018-09-08 \\
    RACE          & 11,949 & 11,938 & 59,690  & 2017-04-15 \\
    SciQ          & 2,996 & 2,996  & 14,980  & 2017-07-19 \\
    \midrule
    \textbf{Total} & \textbf{52,966} & \textbf{52,935} & \textbf{264,761} &  \\
    \bottomrule
    \end{tabular}}
\end{table}

Several well-known benchmarks have been excluded from our study due to their format, task design, or evaluation constraints. Some datasets, such as CRASS \cite{frohberg-binder-2022-crass}, WinoGrande \cite{10.1145/3474381}, PIQA \cite{Bisk2020}, and BoolQ \cite{clark2019boolq}, feature ternary- or binary-choice questions, limiting their applicability to our multiple-choice evaluation framework. Others, like GPQA \cite{rein2024gpqa}, contain highly technical questions with extreme difficulty, making paraphrase evaluation impractical. Benchmarks such as GSM-8K \cite{cobbe2021gsm8k} and MATH \cite{hendrycks2021measuring} require natural language answers assessed via automatic scoring, diverging from our controlled multiple-choice setup. Datasets focused on conversational or multi-turn interactions, including MT-Bench \cite{bai2024mt}, QuAC \cite{choi-etal-2018-quac}, and ACI-Bench \cite{yim2023aci}, were omitted due to their dependency on dialogue history. Similarly, retrieval-based datasets like MS-MARCO \cite{DBLP:journals/corr/NguyenRSGTMD16} and QMSum \cite{zhong2021qmsum} do not align with our evaluation approach, as they emphasize passage retrieval rather than direct question answering. Code-related benchmarks such as BigCodeBench \cite{zhuo2024bigcodebench}, CodeXGLUE \cite{DBLP:journals/corr/abs-2102-04664}, HumanEval \cite{chen2021codex}, and MBPP \cite{austin2021program} were also excluded, as our focus remains on natural language understanding. Lastly, we omitted datasets designed for model-based judgment and ranking, including LLM Judge \cite{zheng2023judging}, Prometheus \cite{kim2023prometheus}, and JudgeLM \cite{zhu2023judgelm}, as they evaluate LLMs in a self-referential manner rather than assessing their robustness to question paraphrasing. 

\begin{table}[tb]
    \centering
    \caption{Four-class accuracy values across benchmarks.}
    \label{tab:model_accuracy_datasets}
    \small
    \adjustbox{max width=\columnwidth}{
    \begin{tabular}{@{}r@{ }lc@{~}c@{~}c@{~}c@{~}c@{~}cc@{}}
\toprule
        & \textbf{Model} & \rotatebox{90}{ARC-C} & \rotatebox{90}{HellaSwag} & \rotatebox{90}{MMLU} & \rotatebox{90}{OpenBookQA} & \rotatebox{90}{RACE} & \rotatebox{90}{SCIQ} & \rotatebox{90}{\textbf{Average}} \\
        \midrule
        1.& Mistral-Large-Instruct-2411 & .94 & .89 & .80 & .88 & .74 & .94 & .87 \\
        2.& Qwen2.5-72B-Instruct & .94 & .83 & .80 & .89 & .74 & .95 & .86 \\
        3.& Llama-3.1-Nemotron-70B-Instruct-HF & .93 & .79 & .81 & .88 & .73 & .95 & .85 \\
        4.& Meta-Llama-3.1-70B-Instruct & .92 & .78 & .80 & .87 & .73 & .95 & .84 \\
        5.& \textit{gpt4omini} & .92 & .82 & .76 & .85 & .71 & .94 & .83 \\
        6.& gemma-2-27b-it & .89 & .77 & .74 & .83 & .70 & .93 & .81 \\
        7.& Qwen2.5-7B-Instruct & .87 & .75 & .70 & .80 & .69 & .92 & .79 \\
        8.& gemma-2-9b-it & .89 & .75 & .71 & .79 & .68 & .91 & .79 \\
        9.& Phi-3-mini-4k-instruct & .85 & .76 & .67 & .79 & .65 & .90 & .77 \\
        10.& Pixtral-12B-2409 & .85 & .74 & .64 & .78 & .67 & .85 & .75 \\
        11.& Mixtral-8x7B-Instruct-v0.1 & .82 & .69 & .68 & .78 & .66 & .89 & .75 \\
        12.& Phi-3-mini-128k-instruct & .83 & .74 & .65 & .76 & .63 & .90 & .75 \\
        13.& aya-expanse-32b & .79 & .75 & .65 & .75 & .67 & .89 & .75 \\
        14.& Mistral-Small-Instruct-2409 & .81 & .61 & .63 & .77 & .63 & .88 & .72 \\
        15.& Nemotron-Mini-4B-Instruct & .78 & .61 & .58 & .76 & .64 & .83 & .70 \\
        16.& Mistral-Nemo-Instruct-2407 & .71 & .53 & .57 & .62 & .61 & .75 & .63 \\
        17.& Meta-Llama-3.1-8B-Instruct & .66 & .43 & .54 & .62 & .57 & .85 & .61 \\
        18.& aya-expanse-8b & .60 & .67 & .53 & .56 & .59 & .72 & .61 \\
        19.& Qwen2-VL-2B-Instruct & .60 & .50 & .49 & .59 & .55 & .79 & .59 \\
        20.& gemma-1.1-7b-it & .64 & .47 & .53 & .54 & .56 & .78 & .59 \\
        21.& zephyr-7b-beta & .63 & .41 & .52 & .55 & .55 & .70 & .56 \\
        22.& Meta-Llama-3-8B-Instruct & .63 & .31 & .51 & .58 & .51 & .80 & .56 \\
        23.& Ministral-8B-Instruct-2410 & .58 & .49 & .50 & .50 & .57 & .65 & .55 \\
        24.& gemma-2-2b-it & .46 & .26 & .38 & .38 & .38 & .70 & .43 \\
        25.& EuroLLM-9B-Instruct & .42 & .26 & .38 & .37 & .41 & .65 & .41 \\
        26.& Mistral-7B-Instruct-v0.1 & .38 & .27 & .38 & .36 & .45 & .55 & .40 \\
        27.& Llama-3.2-1B-Instruct & .33 & .25 & .33 & .43 & .29 & .60 & .37 \\
        28.& gemma-1.1-2b-it & .29 & .26 & .29 & .28 & .29 & .40 & .30 \\
        29.& gemma-7b-it & .22 & .25 & .23 & .27 & .23 & .25 & .24 \\
        30.& falcon-7b-instruct & .22 & .25 & .23 & .27 & .22 & .25 & .24 \\
        31.& Mistral-Nemo-Base-2407 & .22 & .25 & .23 & .27 & .22 & .26 & .24 \\
        32.& bloom-7b1 & .22 & .25 & .23 & .27 & .22 & .25 & .24 \\
        33.& bloom-560m & .22 & .25 & .23 & .27 & .22 & .25 & .24 \\
        34.& EuroLLM-1.7B-Instruct & .22 & .25 & .23 & .27 & .22 & .25 & .24 \\
        \midrule
        & \textbf{Average}  & .63 & .53 & .53 & .59 & .53 & .71 &  \\
        \bottomrule
    \end{tabular}
    }
\end{table}

\subsection{Large Language Models}
We evaluate a diverse selection of 34 LLMs, using a wide range of model sizes, architectures, and training paradigms. They are listed in Table~\ref{tab:model_accuracy_datasets}, which also presents the average effectiveness (accuracy over 4 classes) of LLMs across the 6 benchmarks.
These models include both compact, efficiency-focused variants and large-scale architectures with billions of parameters, allowing us to analyze how model scale influences robustness to paraphrased inputs. The models are predominantly instruction-tuned, meaning they have undergone fine-tuning on human-generated prompts using Reinforcement Learning from Human Feedback (RLHF) \cite{10.5555/3600270.3602281} to improve their ability to follow instructions and generalize across different tasks.
All open-source models in this study have been accessed through Hugging Face, ensuring consistency in implementation and evaluation. Additionally, we include a closed-source reference model, ChatGPT (\texttt{gpt4o-mini}), accessed via OpenAI's API. This allows us to compare publicly available models against a proprietary system that benefits from continuous updates and private optimization strategies.

\subsection{Paraphrases Generation} 
To assess the robustness of LLMs against linguistic variation, we systematically paraphrase benchmark questions while ensuring their semantic integrity remains intact. This allows us to isolate the impact of rewording on model effectiveness, providing a clearer understanding of how linguistic variability affects performance. By maintaining the original intent and reasoning requirements of each question, we ensure that performance changes are due to model robustness rather than altered question difficulty. Additionally, to avoid introducing confounding factors, we preserve the original order of answer choices, ensuring that only the question phrasing is modified.

Given the large scale of our benchmarks, fully manual paraphrasing would be infeasible. 
Therefore, we rely on automatic paraphrasing, which we then validate to control that the generated variations maintain the original meaning and align with the intended answer, while introducing linguistic diversity.
Further details on the validation methodology and results are discussed below in Section~\ref{sec:eva-paraphrases}. Paraphrased versions of the benchmark questions are generated using OpenAI’s \texttt{GPT-4o mini} model,\footnote{\url{https://platform.openai.com/docs/models\#gpt-4o-mini}} accessed via API. The model is prompted to produce five alternative phrasings of each question while adhering to strict constraints to preserve meaning. The paraphrasing prompt is shown in Figure~\ref{fig:paraphrasing_prompt}.
\begin{figure}[tb]
    \centering
    \fbox{
        \parbox{0.95\linewidth}{
        \footnotesize
Paraphrase the following text 5 times, preserving the original meaning and avoiding negations. If it is a question, ensure the paraphrases keep it in question form; if it is a sentence fragment, complete it in a natural way. Each paraphrase must be unique, numbered, and placed on a new line.\\[.5em]
Original Text: \texttt{\{original text\}}.\\
Output each paraphrase as follows:\\
1.\\
2.\\
3.\\
4.\\
5.
}}
\caption{Paraphrasing prompt. \\[2em]}
    \label{fig:paraphrasing_prompt}
\end{figure}

This paraphrasing process allows us to test model robustness to rewording without introducing other changes that could affect performance. While the vast majority of benchmark questions were successfully paraphrased, we found that for a small subset of 31 instances (i.e., less than 0.0004\% of the 53,966 questions, see details in Table~\ref{tab:dataset}), the model refused to generate paraphrases, typically due to the presence of explicit language or content that triggered its safety mechanisms. These few cases were excluded from our experiments to preserve the integrity and consistency of the evaluation.

\subsection{Model Prompting}
To systematically evaluate model responses across benchmarks, we employ a standardized prompting strategy that ensures consistency across all LLMs. Each model is presented with a question in a multiple-choice format and tasked with selecting the most appropriate answer among the four possible options. To maintain uniformity, we enforce constraints on response generation, ensuring that models produce outputs strictly adhering to the predefined answer format of each benchmark.
We use the prompt as in Figure~\ref{fig:evaluation_prompt} to present the questions to the models, where \texttt{\{question\}} is the text of the question, and \texttt{\{choices list\}} refers to the multiple-choice options presented in a numbered list.
The structured format prompt ensures that models generate responses strictly in numerical format, avoids output variability, and allows for straightforward evaluation of correctness. Although we did not systematically experiment with prompt variations, we observed no changes in model responses when modifying the phrasing of the prompt, suggesting that answer selection remained stable across minor prompt rewordings.

For answer selection, we adopt a standard approach based on top-1 token probability, where the model's final prediction corresponds to the highest-probability token in its output distribution. This decoding strategy is widely used in benchmark-based LLM evaluation settings \cite{hendryckstest2021}, as it avoids stochasticity and facilitates reproducibility. Since we preserve the original order of answer choices, the observed performance differences should stem from linguistic variation only, rather than choice position bias.

\begin{figure}[tb]
    \centering
    \fbox{
        \parbox{0.95\linewidth}{
        \footnotesize
You will be asked to answer a question by choosing the correct option number only (1--4).
Please respond with just the number, without any additional text, symbols, or punctuation.\\[.5em]
Question: \texttt{\{question\}}\\
Options: \texttt{\{choices list\}}\\[.5em]
Provide only the number of the correct answer (1--4).
For example, if the correct answer is the first option, respond with \texttt{1} only.\\[.5em]
Answer:
}}
\caption{Evaluation prompt.\\[2em]}
    \label{fig:evaluation_prompt}
\end{figure}

All evaluations are conducted in a zero-shot setting \cite{NEURIPS2022_8bb0d291,wei2021finetuned}, where models receive no examples and are expected to select an answer directly. While this setup avoids prompt engineering, it also limits the model's capacity for structured reasoning. 
To explore the impact of more complex reasoning strategies, we also run some additional experiments. Although we do not have the space to discuss results in detail, we report some data as an indication that they do not seem promising in terms of improving (or changing) the effectiveness and consistency of the models substantially. For example,
we run experiments using Chain-of-Thought (CoT) prompting, in which each model generates 5 reasoning paths per input, followed by majority voting over the predicted answers. 
We found that only a subset of models were able to handle CoT prompting reliably, while others frequently produced hallucinated reasoning steps \cite{10.5555/3600270.3602070,wei2022emergent}. Even among the models that managed CoT effectively, the average improvement in accuracy over the zero-shot setting was typically below $3\%$, with the exception of models 16, 25 and 34, which nonetheless did not achieve accuracies higher than 0.70 (and therefore are not competitive with the best zero-shot models, as can be seen by comparing with the accuracy values in Table~\ref{tab:model_accuracy_datasets}). Moreover, CoT prompting led to a comparable drop in response consistency of approximately $7\%$ across models. 
For these reasons, and to remain aligned with standard benchmark protocols, we focus our evaluation on zero-shot prompting with deterministic top-1 decoding, leaving a deeper investigation into CoT and other complex reasoning strategies for future work.

\section{Results}
We analyze the two research questions in the following subsections.

\subsection{\ref{rq-reliable}: Are Benchmark Evaluations Reliable?}
We discuss how paraphrases affect both answer consistency and effectiveness of the models, as well as paraphrases generation validity.

\subsubsection{Consistency of Models Across Paraphrases}
\begin{figure*}
    \centering
    \includegraphics[width=1\linewidth]{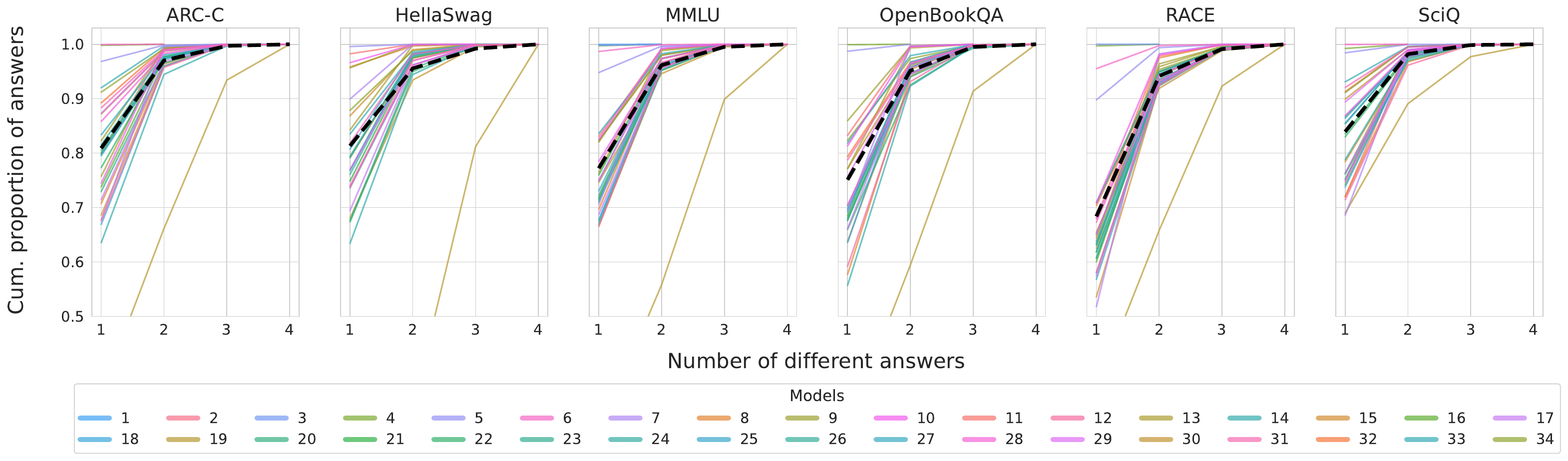}
    \caption{Cumulative proportion of different answers. Model numbers as in Table~\ref{tab:model_accuracy_datasets}.\\[1em]}
    \label{fig:cumsum-answers}
\end{figure*}

We start by analyzing the consistency of model answers across paraphrased versions of the same question. Figure~\ref{fig:cumsum-answers} reports the cumulative distribution of answer diversity across all evaluated models and benchmarks. In each plot (one for each dataset), the x-axis shows the number of different answers (from 1 to 4, as each question has four possible choices) a model gives to paraphrases of the same original question (i.e., 6 possible cases: the original question and the five generated paraphrases). The  y-axis shows the cumulative proportion of such cases. Each colored line corresponds to one model, and the dashed black line denotes the average trend across all models.

The plots show that only a minority of models consistently select the same answer across all paraphrases of a question (the few lines on the top part of the plots, for which the proportion of only one answer given is around $100\%$). Most models vary their answers depending on the specific formulation of the question. 
Even if there is quite a variation across models, we can summarize these data by looking at the average (black dashed line): around $70\%$--$85\%$ of questions receive a single consistent answer from a model, with the remaining $15\%$--$30\%$ of cases showing two, three, or even four distinct answers across paraphrases. 
The fractions of questions receiving more than 2 distinct answers are low but not zero. For the six datasets from left to right, the fractions of questions receiving exactly 3 distinct answers are $2.58\%$, $3.26\%$, $3.31\%$, $3.99\%$, $4.77\%$, $1.6\%$, and 4 distinct answers are $0.26\%$, $0.67\%$, $0.41\%$, $0.41\%$, $0.77\%$, $0.13\%$.
This phenomenon is robust across all benchmarks, with similar patterns observed for all of them.
These results, and specifically the 15\%--30\% of cases for which at least two answers are selected, point to a substantial degree of response variability in current LLMs when evaluated with paraphrased questions provided as inputs. In particular, even state-of-the-art models occasionally exhibit significant sensitivity to surface-level changes in question phrasing. 
While this metric does not directly indicate whether the selected answers are correct, it raises concerns about the stability of benchmark-based evaluations. If a model's output varies across semantically equivalent rewordings, benchmark accuracy scores may reflect the particular wording of test items more than the model's reasoning abilities.

It is important to note that some models (i.e., 1, 2, 3, 5 and 6) demonstrate high consistency, suggesting that more stable behavior is achievable (although often at the expense of low accuracy, as we discuss shortly). However, even the most robust models fail to maintain perfect consistency across all questions.
The plots also show some differences across benchmarks; in particular, models consistently achieve lower consistency on OpenBookQA and, especially, RACE.
These differences might arise from the greater complexity and ambiguity of questions in these datasets. For instance, OpenBookQA and RACE often require multi-step inference.

\subsubsection{Model Accuracy, and Relationship with Consistency}

\begin{figure}[tb]
    \centering
    \includegraphics[width=1\linewidth]{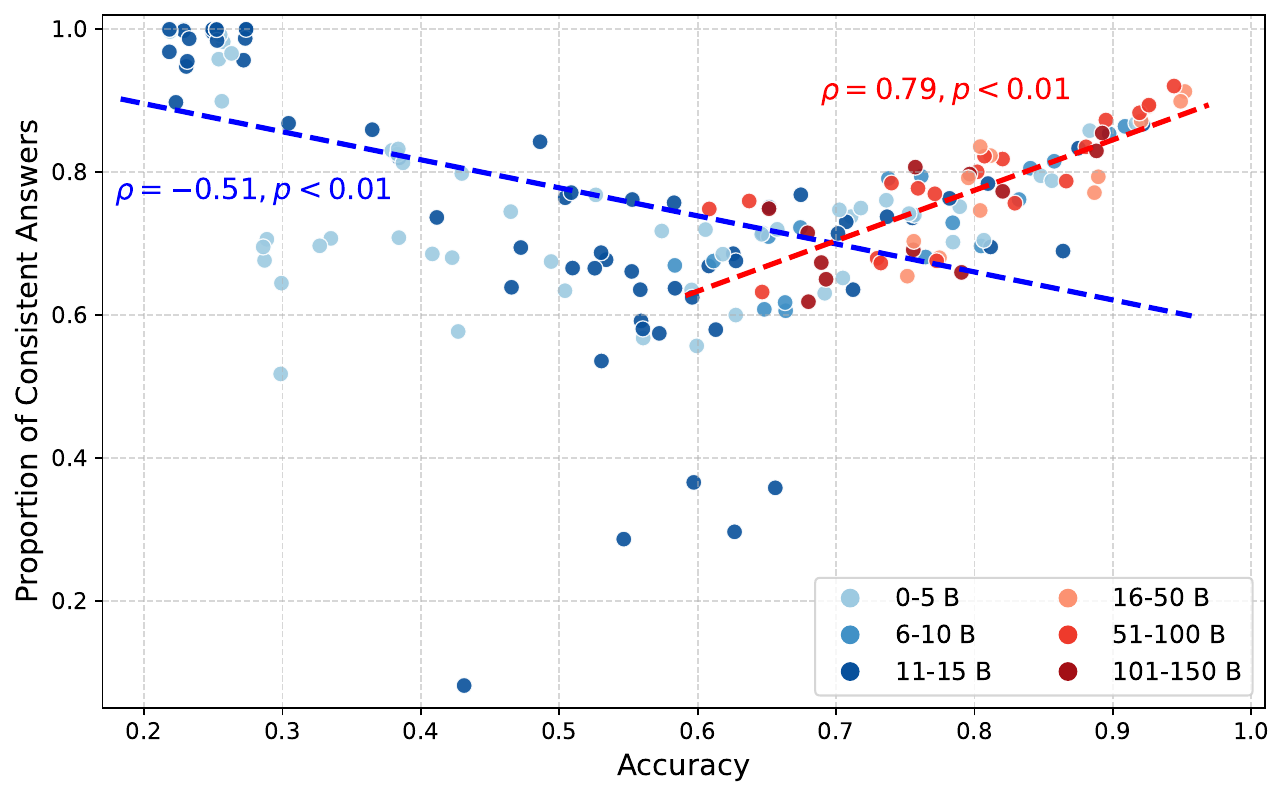}
    \caption{Correlation between models' accuracy and proportion of consistent answers.\\[1em]}
    \label{fig:accuracy-consistency}
\end{figure}

We now turn to investigate whether models that are more consistent are also more accurate, and how effectiveness is affected when question rewordings are introduced. We analyze the correlation between a model's accuracy and its proportion of consistent answers across paraphrased questions over all datasets in Figure~\ref{fig:accuracy-consistency}:  each point represents a model on a dataset, with accuracy on the x-axis and the proportion of consistent answers on the y-axis. Models are divided by size, with smaller models (0--15B parameters) shown in blue and larger models (16--150B) in orange. For each of the two groups, we fit a linear regression line and report the Pearson's $\rho$ correlation coefficient and its associated $p$-value.

As we can see by inspecting the plot and the regression lines, the two model groups exhibit opposite trends. For smaller models, there is a statistically significant negative correlation between accuracy and consistency ($\rho = -0.51$). That is, less capable models tend to be more consistent in their predictions, although they are often wrong. This behavior may be attributed to over-simplicity: such models may repeat the same, often wrong, answer regardless of subtle rewordings due to their limited real understanding of the questions' semantics (i.e., true meaning), although this does not explain why the most ``stubborn'' models are also those that are more frequently wrong.

In contrast, larger models show a strong and significant positive correlation between accuracy and consistency ($\rho = 0.79$), suggesting that as models become larger, they not only improve in accuracy over benchmarks' questions but also become more robust to linguistic variability: these models are more able to generalize and to preserve correct answers across different paraphrased inputs.

These findings confirm what can be perceived from the ``U'' shape of the plot, i.e., that consistency is not always a proxy for correctness: high consistency on paraphrased questions can coincide with low accuracy in under-performing models. However, for more advanced models, consistency becomes a meaningful indicator of robustness and generalization, as they are more likely to preserve correct answers across different questions' phrasings. This duality highlights the importance of interpreting consistency in context, particularly when using it as a proxy for benchmark reliability. 

More generally, our results suggest that benchmark evaluations as often conducted today, i.e., reporting a single average accuracy score over the original questions, may not fully capture a model's true semantic understanding \cite{burnell-etc-2023}. 
Although the models with the highest accuracy (right part of the plot) are also among the most consistent ones, a model with a high average accuracy might in fact be not robust to surface-level changes in question formulation.
This means that benchmarks relying only on original question formulations may overestimate a model's real-world robustness, as they fail to account for performance drops under minor linguistic variations. Consequently, their reliability as tools for LLM evaluation is limited, since true language understanding requires stability in answering across diverse phrasings. Incorporating paraphrased inputs into evaluation protocols can help address this limitation, offering a more realistic and comprehensive measure of model capabilities.

\subsubsection{Evaluation of Generated Paraphrases}
\label{sec:eva-paraphrases}

\begin{figure}[tb]
    \centering
    \includegraphics[width=1\columnwidth]{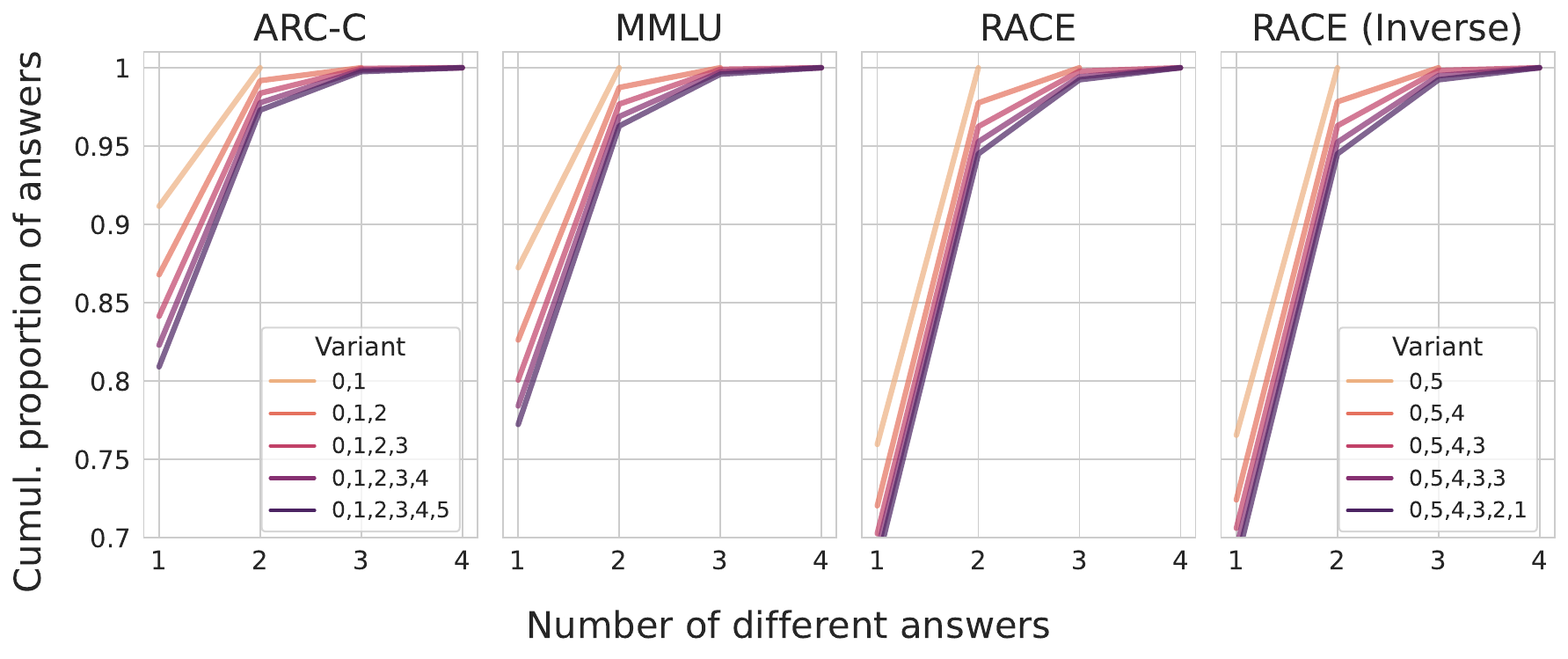}
    \caption{Cumulative proportion of distinct answers as the number of generated paraphrases per question increases, starting from the original question (``0'') plus one generated paraphrase, up to five. In the rightmost plot, the generated paraphrases are in the inverse order.\\[1em]}
    \label{fig:cumsum-answers-para}
\end{figure}

To analyze the reliability of benchmark evaluations, we examine how the number of generated paraphrases influences the consistency of model responses. Figure~\ref{fig:cumsum-answers-para} reports the cumulative proportion of distinct answers produced by models as additional paraphrases are introduced, ranging from only the original question (``0'') to up to six total variants (original plus five paraphrases). Each line represents the average consistency across all models for a given number of available paraphrased versions. Only three benchmarks are shown, as the remaining ones have very similar trends. In the rightmost plot, the paraphrases are presented in reverse order (from the fifth to the first).

By focusing on the first 3 plots of the figure, we observe a clear and consistent trend across all benchmarks: as more paraphrases are added, the number of distinct answers increases. This is reflected by the systematic downward shift of each line as the number of generated paraphrases grows. In particular, the proportion of questions for which models provide a single, consistent answer decreases steadily with each added variant. This means that paraphrasing indeed introduces increasing confusion for LLMs, thereby reducing consistency and revealing their sensitivity to linguistic variation.

The lines are well-separated and ordered as expected, i.e., from ``0'' to ``0,1,2,3,4,5'': this confirms that the generated paraphrases introduce meaningful variability while preserving enough semantic fidelity to challenge the models. If the lines were overlapping, or if the addition of paraphrases did not affect the number of unique responses, it would suggest either low-quality paraphrases (e.g., overly similar or trivially equivalent) or overly rigid models that default to the same answer regardless of wording. Conversely, an inverse or erratic ordering might indicate that paraphrases are not semantically aligned with the original question, thus changing the task rather than preserving it. 
 This interpretation is further supported by the fact that repeating the experiment with paraphrases in reverse order (i.e., ``5,4,3,2,1,0'') produces a nearly identical plot (last plot of Figure~\ref{fig:cumsum-answers-para}), with lines that closely overlap or slightly deviate from the original ones, reinforcing the semantic consistency of the paraphrases.

These results suggest that the paraphrases are effective in injecting linguistic diversity, and that current benchmark practices, typically relying on single question formulations, fail to capture the semantic fragility, or brittleness as discussed by \citet{lewis2024evaluating}, of LLMs. This further motivates the need for robustness-aware evaluation methodologies that account for variation in question wording.

\subsection{\ref{rq-robust}: Are LLMs Robust to Question Paraphrases?}

\begin{figure*}[tb]
    \centering
    \includegraphics[width=1\linewidth]{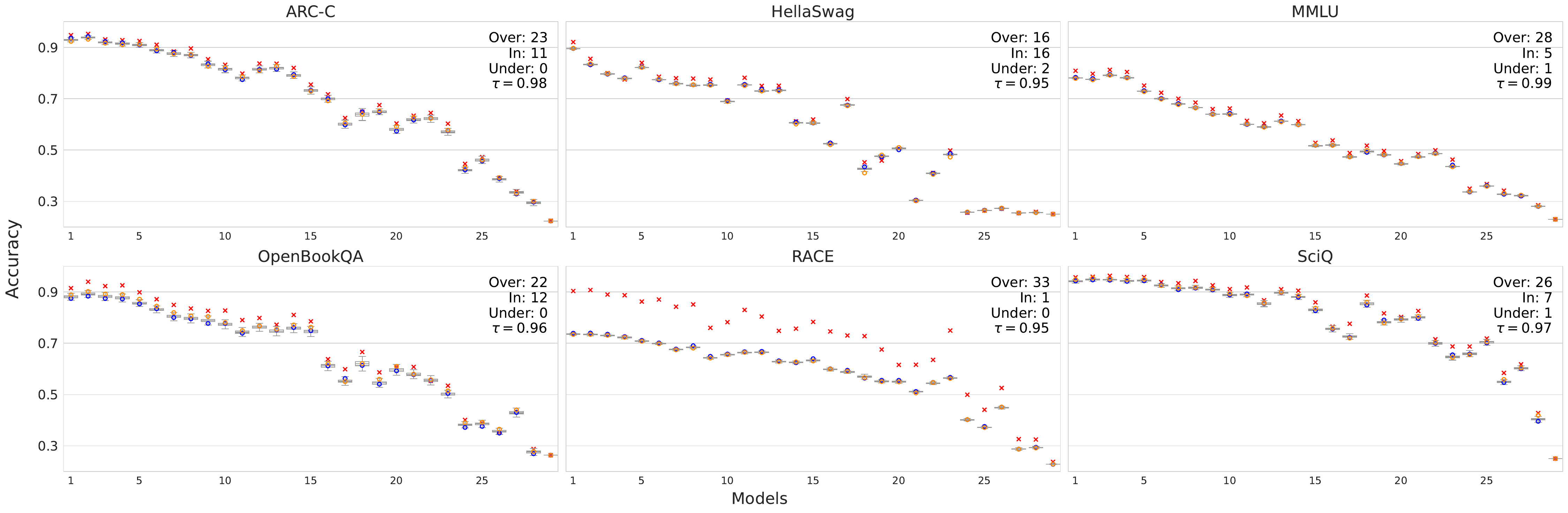}
    \caption{Accuracy on original questions (red crosses), sampled paraphrase sets (boxplots), 1st (orange circle) and 5th (blue circle) generated paraphrases. Models are sorted by original accuracy (Table~\ref{tab:model_accuracy_datasets}; only models 1--29 are shown).\\[1em]}
    \label{fig:rank-robustness}
\end{figure*}

\begin{figure}[tb]
    \centering
    \includegraphics[width=1\linewidth]{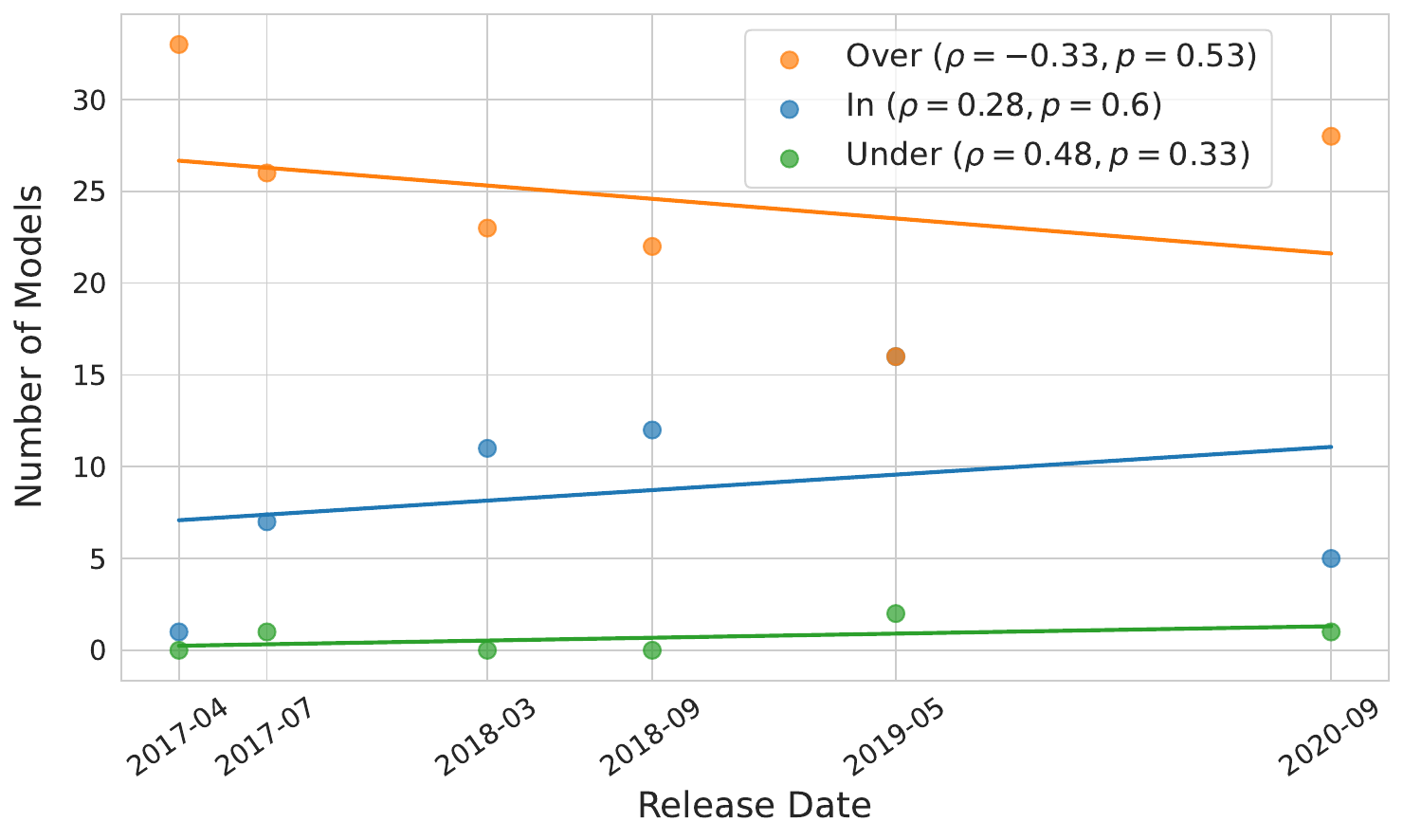}
    \caption{Correlation between the number of models in the \textit{Over}, \textit{In}, and \textit{Under} categories across benchmark release dates (Table \ref{tab:dataset}).\\[1em]}
    \label{fig:rank-robustness-release}
\end{figure}

We now turn to a deeper analysis of whether benchmark-based evaluations remain stable after question paraphrasing. Specifically, we aim to understand if paraphrasing affects how models are ranked on the basis of their accuracy or significantly impacts evaluation outcomes beyond raw accuracy degradation. To this end, Figure~\ref{fig:rank-robustness} presents a multi-faceted view of the relationship between models' effectiveness on original questions and sampled generated paraphrases.

Each plot corresponds to one of the six benchmarks, and shows the performance of models sorted as in Table~\ref{tab:model_accuracy_datasets}, by their average accuracy on the original (i.e., non-paraphrased) benchmarks (i.e., the last column of the table). Only models 1--29 are shown, given the systematically low accuracy of models 30--34.
The red cross indicates the model's accuracy on the original questions for that dataset (i.e., the values in the column of the table corresponding to that dataset), while the gray boxplots represent the distribution of accuracy obtained from 1,000 samples of generated paraphrases. The sampling was performed only on the test set and on the validation set, if present, to focus the evaluation on generalization capabilities. Each sample selects a different random paraphrase for each question, sometimes the original, sometimes one of the five generated variants, thus simulating realistic linguistic variability during inference. This approach allows us to examine how paraphrased question variants affect model accuracy and rankings in aggregate. For example, one sampled instance might use paraphrase 1 for Question 1, paraphrase 3 for Question 2, and so on.

We classify each case into three categories:
(i) \textit{Over}, where the original score is higher than the paraphrase distribution, indicating performance degradation with rewordings; 
(ii) \textit{In}, where the red cross lies inside the interquartile range of the boxplot, suggesting that paraphrasing has minimal impact; 
(iii) \textit{Under}, where the original score is lower than the paraphrase sample accuracy, suggesting that the model benefits from rephrasing.

Across all benchmarks, we observe a clear trend: the majority of models fall into the \textit{Over} category, meaning that their performance degrades when faced with paraphrased inputs. This observation underscores the limited robustness of LLMs to surface-level linguistic changes. Only a small fraction of models remain stable across paraphrases (\textit{In}), and very few show improvements (\textit{Under}). For example, in MMLU, only 5 models fall within the paraphrase boxplot, 28 perform worse, and just 1 improves. 
This pattern is consistent across all benchmarks and it is further supported by the comparison of model accuracy on the first and last generated paraphrase, represented in Figure~\ref{fig:rank-robustness} as orange and blue small circles, respectively. Across benchmarks, these points often alternate in position (i.e., sometimes the first paraphrase yields higher accuracy, sometimes the last) with no consistent pattern. On average, the accuracy across all models is 0.54 for both the first and the last paraphrase, underscoring the semantic equivalence of the variants discussed in Section~\ref{sec:eva-paraphrases}. This result supports the idea that no single paraphrase systematically favors or disadvantages model effectiveness, and thus somehow validates their use for probing model robustness.

We also compute Kendall's $\tau$ between the rankings derived from original accuracy and those derived from median paraphrased performance (i.e., the central horizontal line of each boxplot in figure, for each dataset). The resulting $\tau$ values are shown in figure and are all above $0.9$ and statistically significant ($p<0.01$): they indicate that despite accuracy fluctuations, the relative ordering of models remains largely preserved. This is a quantitative confirmation of the trends that can be intuitively seen in figure.

Finally, the plot in Figure~\ref{fig:rank-robustness-release} explores the relationship between benchmark release date and the number of models falling into each of the three accuracy-related categories (\textit{Over}, \textit{In}, \textit{Under}). We observe a not-significant negative correlation for the \textit{Over} category ($\rho = -0.33$): for older benchmarks there is the tendency to have more models whose accuracy on the original questions is higher than on paraphrased ones. Conversely, the \textit{In} category shows a weak positive correlation ($\rho = 0.28$), indicating that newer benchmarks tend to yield more stable results across paraphrases. Taken together, these trends suggest the conjecture that models may overfit to older benchmarks, potentially due to data contamination from pretraining, thus achieving inflated accuracy on original formulations. It seems that paraphrasing disrupts this memorization/leak effect, exposing weaker generalization. On the other hand, more recent benchmarks appear less affected, either because they are less likely to be part of training data or because they are designed to better resist shallow memorization. The \textit{Under} category shows a positive trend ($\rho = 0.48$), opposite to that for \textit{Over}, consistently with the above conjecture (although the absolute numbers are very low).

In summary, these findings reveal that while model rankings are largely preserved, paraphrasing exposes serious limitations in model generalization, and benchmarks may overestimate true model capabilities.

\section{Conclusions and Future Work}
Our study reveals insights into the robustness of LLMs to linguistic variability and the reliability of benchmark-based evaluations. We show that while model rankings tend to remain relatively stable when benchmark questions are paraphrased, absolute accuracy scores drop significantly. This discrepancy suggests that current evaluations, based on static question wordings, may overestimate models' true generalization capabilities by failing to account for natural linguistic variation. Our findings highlight the need for evaluation methodologies that better reflect real-world usage.

We also show that consistency across paraphrased inputs (i.e., providing the same answer to different formulations of a question) is not always a reliable indicator of correctness. Less capable models can exhibit high consistency while consistently providing wrong answers, whereas stronger models tend to be both consistent and accurate. This suggests the necessity of interpreting consistency metrics carefully and discourages relying solely on average benchmark scores as a proxy for reasoning ability. 
We also uncover some evidence of potential data contamination, particularly in older benchmarks, where models show lower agreement between original and paraphrased performances. This raises concerns about memorization rather than genuine understanding, emphasizing the importance of evaluating models on fresh, unseen, and linguistically diverse inputs.

Our work has limitations, that call for future developments. 
We addressed exclusively multiple-choice benchmarks and left out open-ended tasks, which may exhibit different patterns. 
While paraphrases were automatically generated and validated to ensure semantic fidelity, they may not perfectly capture the full spectrum of real-world linguistic variability. Further validation of the generated paraphrases is needed. We plan to do so with manual checks  and by using alternative generation processes. Indeed, we plan to explore paraphrasing strategies that better mimic natural human linguistic diversity, including user-generated ones. We also plan to extend our analysis to more complex reasoning settings and to release paraphrase-augmented benchmark suites. Ultimately, we advocate for a shift from static, rigid benchmarks towards dynamic, linguistically diverse evaluation frameworks that better capture the complexities of real-world language.

\clearpage
\paragraph*{Acknowledgments}
This research is partially supported by the PRIN 2022 Project – ``MoT—The Measure of Truth: An Evaluation-Centered Machine-Human Hybrid Framework for Assessing Information Truthfulness'' – Code No. 20227F2ZN3, CUP No. G53D23002800006 Funded by the European Union – Next Generation EU – PNRR M4 C2 I1.1. and by the ESF+ 2021/2027 Regional Program of the Autonomous Region Friuli Venezia Giulia (PPO 2023, Program No. 22/23 – LINE A: PhD programmes).

\bibliography{custom}

\begin{thebibliography}{51}
\providecommand{\natexlab}[1]{#1}
\providecommand{\url}[1]{\texttt{#1}}
\expandafter\ifx\csname urlstyle\endcsname\relax
  \providecommand{\doi}[1]{doi: #1}\else
  \providecommand{\doi}{doi: \begingroup \urlstyle{rm}\Url}\fi

\bibitem[Austin et~al.(2021)]{austin2021program}
J.~Austin et~al.
\newblock Program synthesis with large language models.
\newblock \emph{arXiv:2108.07732}, 2021.

\bibitem[Bai et~al.(2024)]{bai2024mt}
G.~Bai et~al.
\newblock Mt-bench-101: A fine-grained benchmark for evaluating large language models in multi-turn dialogues.
\newblock \emph{arXiv:2402.14762}, 2024.

\bibitem[Bailey et~al.(2008)Bailey, Craswell, Soboroff, Thomas, de~Vries, and Yilmaz]{10.1145/1390334.1390447}
P.~Bailey, N.~Craswell, I.~Soboroff, P.~Thomas, A.~P. de~Vries, and E.~Yilmaz.
\newblock Relevance assessment: are judges exchangeable and does it matter.
\newblock In \emph{Proc. of SIGIR}, page 667–674, 2008.

\bibitem[Bisk et~al.(2020)Bisk, Zellers, Bras, Gao, and Choi]{Bisk2020}
Y.~Bisk, R.~Zellers, R.~L. Bras, J.~Gao, and Y.~Choi.
\newblock Piqa: Reasoning about physical commonsense in natural language.
\newblock In \emph{AAAI}, 2020.

\bibitem[Burnell et~al.(2023)]{burnell-etc-2023}
R.~Burnell et~al.
\newblock {Rethink reporting of evaluation results in AI}.
\newblock \emph{Science}, 380\penalty0 (6641):\penalty0 136--138, 2023.

\bibitem[Carterette et~al.(2006)Carterette, Allan, and Sitaraman]{10.1145/1148170.1148219}
B.~Carterette, J.~Allan, and R.~Sitaraman.
\newblock Minimal test collections for retrieval evaluation.
\newblock In \emph{Proc. of SIGIR}, page 268–275, 2006.

\bibitem[Carterette et~al.(2015)Carterette, Bah, and Zengin]{10.1145/2808194.2809470}
B.~Carterette, A.~Bah, and M.~Zengin.
\newblock Dynamic test collections for retrieval evaluation.
\newblock In \emph{Proc. of the 2015 International Conference on The Theory of Information Retrieval}, page 91–100, 2015.

\bibitem[Chen et~al.(2021)]{chen2021codex}
M.~Chen et~al.
\newblock Evaluating large language models trained on code.
\newblock \emph{arXiv:2107.03374}, 2021.

\bibitem[Choi et~al.(2018)Choi, He, Iyyer, Yatskar, Yih, Choi, Liang, and Zettlemoyer]{choi-etal-2018-quac}
E.~Choi, H.~He, M.~Iyyer, M.~Yatskar, W.-t. Yih, Y.~Choi, P.~Liang, and L.~Zettlemoyer.
\newblock {Q}u{AC}: Question answering in context.
\newblock In \emph{Proc. of the 2018 EMNLP}, pages 2174--2184, 2018.

\bibitem[Clark et~al.(2019)Clark, Lee, Chang, Kwiatkowski, Collins, and Toutanova]{clark2019boolq}
C.~Clark, K.~Lee, M.-W. Chang, T.~Kwiatkowski, M.~Collins, and K.~Toutanova.
\newblock Boolq: Exploring the surprising difficulty of natural yes/no questions.
\newblock In \emph{NAACL}, 2019.

\bibitem[Clark et~al.(2018)Clark, Cowhey, Etzioni, Khot, Sabharwal, Schoenick, and Tafjord]{clark2018think}
P.~Clark, I.~Cowhey, O.~Etzioni, T.~Khot, A.~Sabharwal, C.~Schoenick, and O.~Tafjord.
\newblock {Think you have Solved Question Answering? Try ARC, the AI2 Reasoning Challenge}.
\newblock \emph{arXiv:1803.05457}, 2018.

\bibitem[Cobbe et~al.(2021)Cobbe, Kosaraju, Bavarian, Chen, Jun, Kaiser, Plappert, Tworek, Hilton, Nakano, Hesse, and Schulman]{cobbe2021gsm8k}
K.~Cobbe, V.~Kosaraju, M.~Bavarian, M.~Chen, H.~Jun, L.~Kaiser, M.~Plappert, J.~Tworek, J.~Hilton, R.~Nakano, C.~Hesse, and J.~Schulman.
\newblock Training verifiers to solve math word problems.
\newblock \emph{arXiv:2110.14168}, 2021.

\bibitem[Frohberg and Binder(2022)]{frohberg-binder-2022-crass}
J.~Frohberg and F.~Binder.
\newblock {CRASS: A Novel Data Set and Benchmark to Test Counterfactual Reasoning of Large Language Models}.
\newblock In \emph{Proc. of 13th LREC}, pages 2126--2140, 2022.

\bibitem[Guiver et~al.(2009)Guiver, Mizzaro, and Robertson]{10.1145/1629096.1629099}
J.~Guiver, S.~Mizzaro, and S.~Robertson.
\newblock A few good topics: Experiments in topic set reduction for retrieval evaluation.
\newblock \emph{TOIS}, 27\penalty0 (4), 2009.

\bibitem[Harman(2011)]{harman2011information}
D.~Harman.
\newblock \emph{Information retrieval evaluation}.
\newblock Morgan \& Claypool, 2011.

\bibitem[Hendrycks et~al.(2021{\natexlab{a}})Hendrycks, Burns, Basart, Zou, Mazeika, Song, and Steinhardt]{hendryckstest2021}
D.~Hendrycks, C.~Burns, S.~Basart, A.~Zou, M.~Mazeika, D.~Song, and J.~Steinhardt.
\newblock {Measuring Massive Multitask Language Understanding}.
\newblock \emph{ICLR}, 2021{\natexlab{a}}.

\bibitem[Hendrycks et~al.(2021{\natexlab{b}})]{hendrycks2021measuring}
D.~Hendrycks et~al.
\newblock Measuring mathematical problem solving with the math dataset.
\newblock \emph{arXiv:2103.03874}, 2021{\natexlab{b}}.

\bibitem[Kim et~al.(2023)]{kim2023prometheus}
S.~Kim et~al.
\newblock Prometheus: Inducing fine-grained evaluation capability in language models.
\newblock In \emph{ICLR}, 2023.

\bibitem[Kojima et~al.(2022)Kojima, Gu, Reid, Matsuo, and Iwasawa]{NEURIPS2022_8bb0d291}
T.~Kojima, S.~S. Gu, M.~Reid, Y.~Matsuo, and Y.~Iwasawa.
\newblock {Large Language Models are Zero-Shot Reasoners}.
\newblock In \emph{NeurIPS}, volume~35, pages 22199--22213, 2022.

\bibitem[Lai et~al.(2017)Lai, Xie, Liu, Yang, and Hovy]{lai-etal-2017-race}
G.~Lai, Q.~Xie, H.~Liu, Y.~Yang, and E.~Hovy.
\newblock {RACE: Large-scale {R}e{A}ding Comprehension Dataset From Examinations}.
\newblock In \emph{Proc. of the 2017 EMNLP}, pages 785--794, 2017.

\bibitem[Lewis and Mitchell(2024)]{lewis2024evaluating}
M.~Lewis and M.~Mitchell.
\newblock Evaluating the robustness of analogical reasoning in large language models.
\newblock \emph{arXiv:2411.14215}, 2024.

\bibitem[Lu et~al.(2021)Lu, Guo, Ren, Huang, Svyatkovskiy, Blanco, Clement, Drain, Jiang, Tang, Li, Zhou, Shou, Zhou, Tufano, Gong, Zhou, Duan, Sundaresan, Deng, Fu, and Liu]{DBLP:journals/corr/abs-2102-04664}
S.~Lu, D.~Guo, S.~Ren, J.~Huang, A.~Svyatkovskiy, A.~Blanco, C.~B. Clement, D.~Drain, D.~Jiang, D.~Tang, G.~Li, L.~Zhou, L.~Shou, L.~Zhou, M.~Tufano, M.~Gong, M.~Zhou, N.~Duan, N.~Sundaresan, S.~K. Deng, S.~Fu, and S.~Liu.
\newblock Codexglue: {A} machine learning benchmark dataset for code understanding and generation.
\newblock \emph{CoRR}, abs/2102.04664, 2021.

\bibitem[Lunardi et~al.(2024)Lunardi, La~Barbera, and Roitero]{10.1145/3627673.3680002}
R.~Lunardi, D.~La~Barbera, and K.~Roitero.
\newblock {The Elusiveness of Detecting Political Bias in Language Models}.
\newblock In \emph{Proc. 33rd ACM Int. Conf. on Information and Knowledge Management}, page 3922–3926, 2024.

\bibitem[Mihaylov et~al.(2018)Mihaylov, Clark, Khot, and Sabharwal]{mihaylov-etal-2018-suit}
T.~Mihaylov, P.~Clark, T.~Khot, and A.~Sabharwal.
\newblock {Can a Suit of Armor Conduct Electricity? A New Dataset for Open Book Question Answering}.
\newblock In \emph{Proc. of the 2018 EMNLP}, pages 2381--2391, 2018.

\bibitem[Mitchell(2023{\natexlab{a}})]{doi:10.1126/science.adj5957}
M.~Mitchell.
\newblock {How do we know how smart AI systems are?}
\newblock \emph{Science}, 381\penalty0 (6654):\penalty0 eadj5957, 2023{\natexlab{a}}.

\bibitem[Mitchell(2023{\natexlab{b}})]{mitchell2023abstraction}
M.~Mitchell.
\newblock {Abstraction and analogy in AI}.
\newblock \emph{Annals of the New York Academy of Sciences}, 1524\penalty0 (1):\penalty0 17--21, 2023{\natexlab{b}}.

\bibitem[Nguyen et~al.(2016)Nguyen, Rosenberg, Song, Gao, Tiwary, Majumder, and Deng]{DBLP:journals/corr/NguyenRSGTMD16}
T.~Nguyen, M.~Rosenberg, X.~Song, J.~Gao, S.~Tiwary, R.~Majumder, and L.~Deng.
\newblock {MS} {MARCO:} {A} human generated machine reading comprehension dataset.
\newblock \emph{CoRR}, abs/1611.09268, 2016.

\bibitem[Ouyang et~al.(2024)Ouyang, Wu, Jiang, Almeida, Wainwright, Mishkin, Zhang, Agarwal, Slama, Ray, Schulman, Hilton, Kelton, Miller, Simens, Askell, Welinder, Christiano, Leike, and Lowe]{10.5555/3600270.3602281}
L.~Ouyang, J.~Wu, X.~Jiang, D.~Almeida, C.~L. Wainwright, P.~Mishkin, C.~Zhang, S.~Agarwal, K.~Slama, A.~Ray, J.~Schulman, J.~Hilton, F.~Kelton, L.~Miller, M.~Simens, A.~Askell, P.~Welinder, P.~Christiano, J.~Leike, and R.~Lowe.
\newblock Training language models to follow instructions with human feedback.
\newblock In \emph{Proc. of NIPS}, 2024.

\bibitem[Parry et~al.(2025)Parry, Fr{\"o}be, Scells, Schlatt, Faggioli, Zerhoudi, MacAvaney, and Yang]{parry2025variations}
A.~Parry, M.~Fr{\"o}be, H.~Scells, F.~Schlatt, G.~Faggioli, S.~Zerhoudi, S.~MacAvaney, and E.~Yang.
\newblock Variations in relevance judgments and the shelf life of test collections.
\newblock \emph{arXiv:2502.20937}, 2025.

\bibitem[Rein et~al.(2024)Rein, Hou, Stickland, Petty, Pang, Dirani, Michael, and Bowman]{rein2024gpqa}
D.~Rein, B.~L. Hou, A.~C. Stickland, J.~Petty, R.~Y. Pang, J.~Dirani, J.~Michael, and S.~R. Bowman.
\newblock {GPQA: A Graduate-Level Google-Proof Q\&A Benchmark}.
\newblock In \emph{First Conf. on Language Modeling}, 2024.

\bibitem[Reuel-Lamparth et~al.(2024)Reuel-Lamparth, Hardy, Smith, Lamparth, Hardy, and Kochenderfer]{reuel2024betterbench}
A.~Reuel-Lamparth, A.~Hardy, C.~Smith, M.~Lamparth, M.~Hardy, and M.~J. Kochenderfer.
\newblock {BetterBench: Assessing AI Benchmarks, Uncovering Issues, and Establishing Best Practices}.
\newblock \emph{Advances in Neural Information Processing Systems}, 37:\penalty0 21763--21813, 2024.

\bibitem[Sakaguchi et~al.(2021)Sakaguchi, Bras, Bhagavatula, and Choi]{10.1145/3474381}
K.~Sakaguchi, R.~L. Bras, C.~Bhagavatula, and Y.~Choi.
\newblock {WinoGrande: An Adversarial Winograd Schema Challenge at Scale}.
\newblock \emph{Commun. ACM}, 64\penalty0 (9):\penalty0 99–106, 2021.

\bibitem[Salido et~al.(2025)Salido, Gonzalo, and Marco]{salido2025none}
E.~S. Salido, J.~Gonzalo, and G.~Marco.
\newblock {None of the Others: a General Technique to Distinguish Reasoning from Memorization in Multiple-Choice LLM Evaluation Benchmarks}.
\newblock \emph{arXiv:2502.12896}, 2025.

\bibitem[Sanderson(2010)]{INR-009}
M.~Sanderson.
\newblock Test collection based evaluation of information retrieval systems.
\newblock \emph{Foundations and Trends in IR}, 4\penalty0 (4):\penalty0 247--375, 2010.

\bibitem[Sclar et~al.(2024)Sclar, Choi, Tsvetkov, and Suhr]{sclar2024quantifying}
M.~Sclar, Y.~Choi, Y.~Tsvetkov, and A.~Suhr.
\newblock {Quantifying Language Models' Sensitivity to Spurious Features in Prompt Design or: How I learned to start worrying about prompt formatting}.
\newblock In \emph{The Twelfth International Conference on Learning Representations}, 2024.

\bibitem[Sparck~Jones and Van~Rijsbergen(1976)]{sparck1976information}
K.~Sparck~Jones and C.~J. Van~Rijsbergen.
\newblock Information retrieval test collections.
\newblock \emph{Journal of documentation}, 32\penalty0 (1):\penalty0 59--75, 1976.

\bibitem[Stevenson et~al.(2024)Stevenson, Pafford, van~der Maas, and Mitchell]{stevenson2024can}
C.~E. Stevenson, A.~Pafford, H.~L. van~der Maas, and M.~Mitchell.
\newblock Can large language models generalize analogy solving like people can?
\newblock \emph{arXiv:2411.02348}, 2024.

\bibitem[Voorhees(1998)]{voorhees1998variations}
E.~M. Voorhees.
\newblock Variations in relevance judgments and the measurement of retrieval effectiveness.
\newblock In \emph{Proc. of SIGIR}, pages 315--323, 1998.

\bibitem[Voorhees(2001{\natexlab{a}})]{10.1145/383952.383963}
E.~M. Voorhees.
\newblock Evaluation by highly relevant documents.
\newblock In \emph{Proc. 24th SIGIR}, page 74–82, 2001{\natexlab{a}}.

\bibitem[Voorhees(2001{\natexlab{b}})]{voorhees2001philosophy}
E.~M. Voorhees.
\newblock The philosophy of information retrieval evaluation.
\newblock In \emph{Workshop of the cross-language evaluation forum for european languages}, pages 355--370, 2001{\natexlab{b}}.

\bibitem[Wei et~al.(2021)Wei, Bosma, Zhao, Guu, Yu, Lester, Du, Dai, and Le]{wei2021finetuned}
J.~Wei, M.~Bosma, V.~Y. Zhao, K.~Guu, A.~W. Yu, B.~Lester, N.~Du, A.~M. Dai, and Q.~V. Le.
\newblock {Finetuned language models are zero-shot learners}.
\newblock \emph{arXiv:2109.01652}, 2021.

\bibitem[Wei et~al.(2022{\natexlab{a}})Wei, Wang, Schuurmans, Bosma, Ichter, Xia, Chi, Le, and Zhou]{10.5555/3600270.3602070}
J.~Wei, X.~Wang, D.~Schuurmans, M.~Bosma, B.~Ichter, F.~Xia, E.~H. Chi, Q.~V. Le, and D.~Zhou.
\newblock {Chain-of-thought prompting elicits reasoning in large language models}.
\newblock In \emph{Proc. of the 36th NeurIPS}, 2022{\natexlab{a}}.

\bibitem[Wei et~al.(2022{\natexlab{b}})]{wei2022emergent}
J.~Wei et~al.
\newblock Emergent abilities of large language models.
\newblock \emph{arXiv preprint arXiv:2206.07682}, 2022{\natexlab{b}}.

\bibitem[Welbl et~al.(2017)Welbl, Liu, and Gardner]{welbl-etal-2017-crowdsourcing}
J.~Welbl, N.~F. Liu, and M.~Gardner.
\newblock {Crowdsourcing Multiple Choice Science Questions}.
\newblock In \emph{Proc. of the 3rd Workshop on Noisy User-generated Text}, pages 94--106, 2017.

\bibitem[Yim et~al.(2023)Yim, Fu, Ben~Abacha, Snider, Lin, and Yetisgen]{yim2023aci}
W.-w. Yim, Y.~Fu, A.~Ben~Abacha, N.~Snider, T.~Lin, and M.~Yetisgen.
\newblock Aci-bench: a novel ambient clinical intelligence dataset for benchmarking automatic visit note generation.
\newblock \emph{Scientific data}, 10\penalty0 (1):\penalty0 586, 2023.

\bibitem[Zellers et~al.(2019)Zellers, Holtzman, Bisk, Farhadi, and Choi]{zellers-etal-2019-hellaswag}
R.~Zellers, A.~Holtzman, Y.~Bisk, A.~Farhadi, and Y.~Choi.
\newblock {H}ella{S}wag: Can a machine really finish your sentence?
\newblock In \emph{Proc. of the 57th ACL}, pages 4791--4800, 2019.

\bibitem[Zhao et~al.(2024)Zhao, Yan, Sun, Xing, Wang, Meng, Cheng, Ren, and Yin]{zhao-etal-2024-improving}
Y.~Zhao, L.~Yan, W.~Sun, G.~Xing, S.~Wang, C.~Meng, Z.~Cheng, Z.~Ren, and D.~Yin.
\newblock {Improving the Robustness of Large Language Models via Consistency Alignment}.
\newblock In \emph{LREC-COLING}, pages 8931--8941, 2024.

\bibitem[Zheng et~al.(2023)]{zheng2023judging}
L.~Zheng et~al.
\newblock {Judging LLM-as-a-Judge with MT-Bench and Chatbot Arena}.
\newblock \emph{Advances in NeurIPS}, 36:\penalty0 46595--46623, 2023.

\bibitem[Zhong et~al.(2021)]{zhong2021qmsum}
M.~Zhong et~al.
\newblock {QMSum: A new benchmark for query-based multi-domain meeting summarization}.
\newblock \emph{arXiv:2104.05938}, 2021.

\bibitem[Zhu et~al.(2023)Zhu, Wang, and Wang]{zhu2023judgelm}
L.~Zhu, X.~Wang, and X.~Wang.
\newblock {JudgeLM: Fine-tuned Large Language Models are Scalable Judges}.
\newblock \emph{arXiv:2310.17631}, 2023.

\bibitem[Zhuo et~al.(2024)]{zhuo2024bigcodebench}
T.~Y. Zhuo et~al.
\newblock Bigcodebench: Benchmarking code generation with diverse function calls and complex instructions.
\newblock \emph{arXiv:2406.15877}, 2024.

\end{thebibliography}
\end{document}